\documentclass[letterpaper, 10 pt, conference]{ieeeconf}  %

\IEEEoverridecommandlockouts                              %

\usepackage{graphicx} %
\usepackage{array,booktabs,arydshln}

\usepackage{cite}
\usepackage{algorithm}
\usepackage{algorithmic}
\usepackage{xcolor}

\definecolor{green}{RGB}{11,155,13}

\usepackage{amsmath}
\usepackage{bbm}
\usepackage[algo2e, ruled, linesnumbered]{algorithm2e}
\DeclareMathOperator*{\argmin}{arg\,min}
\usepackage{multirow}

\usepackage{etoolbox}
\usepackage{dsfont}
\makeatletter
\patchcmd{\@makecaption}
  {\scshape}
  {}
  {}
  {}
\makeatother

\title{\LARGE \bf
APPLI: Adaptive Planner Parameter Learning
From Interventions
}

\author{Zizhao Wang$^{*1}$, Xuesu Xiao$^{*2}$, Bo Liu$^{*2}$, Garrett Warnell$^{2,3}$, and Peter Stone$^{2,4}$%
\thanks{$^*$Equal Contribution}%
\thanks{$^{1}$Department of Electrical and Computer Engineering,
        University of Texas at Austin, Austin, Texas 78712
        {\tt\small zizhao.wang@utexas.edu}}%
\thanks{$^{2}$Department of Computer Science,
        University of Texas at Austin, Austin, Texas 78712
        {\tt\small \{xiao, bliu, pstone\}@cs.utexas.edu}}%
\thanks{$^{3}$Computational and Information Sciences Directorate, Army Research Laboratory, Austin, Texas 78712
        {\tt\small garrett.a.warnell.civ@mail.mil}}%
\thanks{$^{4}$Sony AI
        }%
}

\usepackage{fancyhdr}

\begin{document}
\maketitle
\thispagestyle{fancy}

\begin{abstract}
While classical autonomous navigation systems can typically move robots from one point to another safely and in a collision-free manner, these systems may fail or produce suboptimal behavior in certain scenarios.
The current practice in such scenarios is to manually re-tune the system's parameters, e.g. max speed, sampling rate, inflation radius, to optimize performance. This practice requires expert knowledge and may jeopardize performance in the originally good scenarios.
Meanwhile, it is relatively easy for a human to identify those failure or suboptimal cases and provide a teleoperated \emph{intervention} to correct the failure or suboptimal behavior. In this work, we seek to learn from those human interventions to improve navigation performance. In particular, we propose \emph{Adaptive Planner Parameter Learning from Interventions} (\textsc{appli}), in which multiple sets of navigation parameters are learned during training and applied based on a confidence measure to the underlying navigation system during deployment. In our physical experiments, the robot achieves better performance compared to the planner with static default parameters, and even dynamic parameters learned from a full human demonstration. We also show \textsc{appli}'s generalizability in another unseen physical test course, and a suite of 300 simulated navigation environments.
\end{abstract}

\section{INTRODUCTION}
\label{sec::intro}

Decades of research has been devoted to developing mobile robot navigation systems that are capable of moving a robot safely from one point to another in obstacle-occupied spaces without collisions. 
Classical navigation systems, such as Elastic-Bands~\cite{quinlan1993elastic} or Dynamic Window Approach (\textsc{DWA})~\cite{fox1997dynamic}, have been robustly deployed on mobile robots with verifiable guarantees of safety and explainability and are able to achieve optimal navigation in most cases. 

\begin{figure}
  \centering
  \includegraphics[width=\columnwidth]{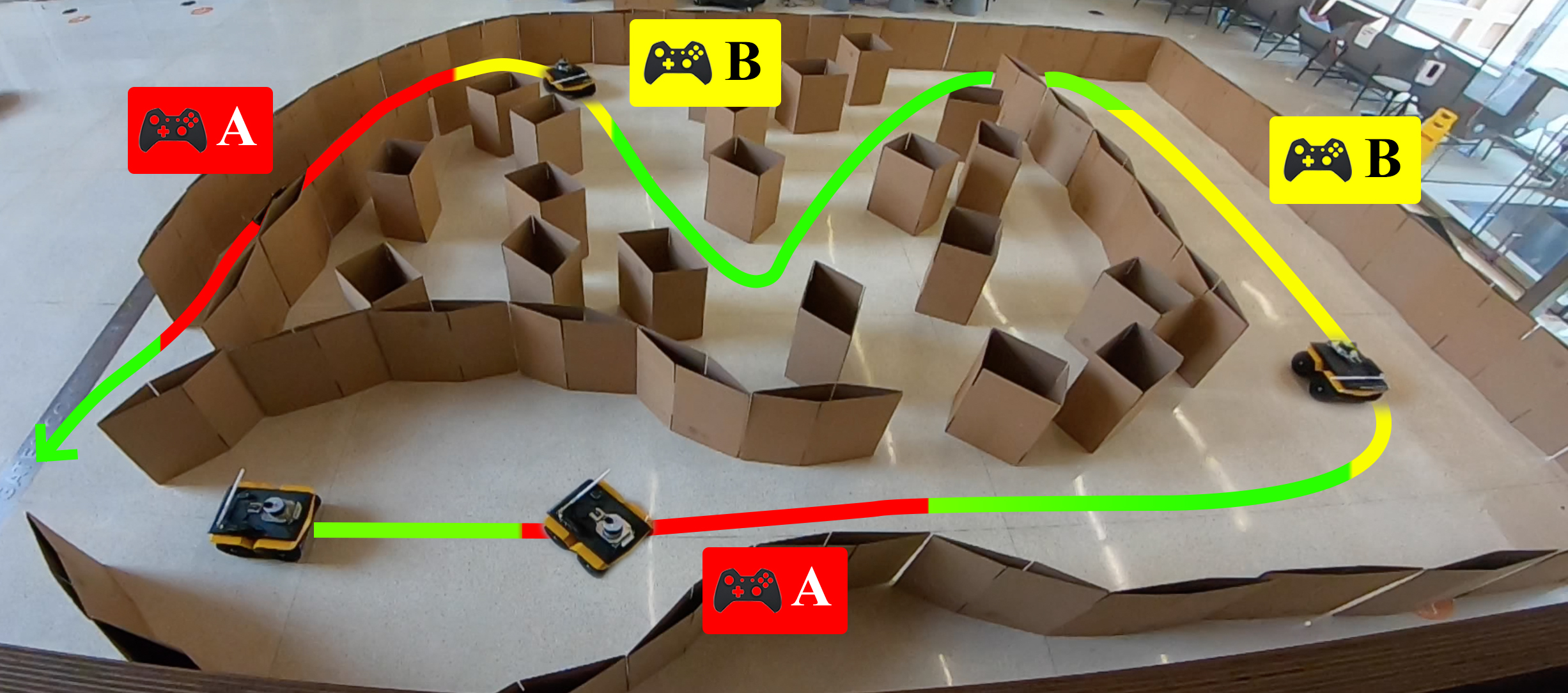}
  \caption{While classical navigation systems perform well in most places (green), they may fail (red) or suffer from suboptimal behavior (yellow) in others. \textsc{appli} utilizes human interventions in these two scenarios (we name them Type A and Type B interventions, respectively) to learn adaptive planner parameters and, based on a confidence measure, uses them during deployment. }
  \label{fig::appli}
\end{figure}

However, in some situations, those classical navigation systems fail or suffer from suboptimal behaviors (Fig. \ref{fig::appli}). For example, the robot may not be able to find a feasible action in highly-constrained spaces~\cite{xiao2020toward, liu2020lifelong}, or may drive unnecessarily slowly in open spaces~\cite{xiao2020agile}. The current solution to these problems is to manually re-tune the parameters of the underlying navigation system (e.g. max speed, sampling rate, inflation radius) to correct the failure cases or suboptimal behaviors in those places. This re-tuning process not only requires expert knowledge onsite during deployment, but also runs the risk of the re-tuned parameters targeted at the failed or suboptimal comprising performance in the originally good scenarios. 

Meanwhile, even a non-expert user (i.e., one who is not familiar with the inner workings of the underlying navigation system) can easily identify the situations where the robot fails or performs suboptimally by watching, and then can intervene by teleoperating the vehicle. In this work, we utilize those human interventions to improve future autonomous navigation in those troublesome places, while maintaining good performance in others. 

In particular, we introduce \emph{Adaptive Planner Parameter Learning from Interventions} (\textsc{appli}). With a set of teleoperated human interventions, \textsc{appli} learns a set of navigation parameters, which are selected dynamically to eliminate failures or suboptimal behaviors during deployment. To assure the learned parameters will not jeopardize navigation performance in other places, the robot only uses the learned parameters when it is confident that they will benefit the current navigation. In our experiments, \textsc{appli} learns from interventions in a real-world navigation task. We test \textsc{appli} in the same training and another unseen physical environment. More than twenty thousands simulation trials are conducted in unseen environments to test \textsc{appli}'s generalizability. Our results show \textsc{appli} can improve upon default navigation performance and can generalize well to unseen environments, indicating that interventions are a uniquely valuable form of human interaction for building navigation systems.

\section{RELATED WORK}
\label{sec::related}

In this section, we review existing work on adaptive planner parameters, learning from intervention, and uncertainty measurement in deep learning. 

\subsection{Adaptive Parameters for Classical Navigation} 
Classical navigation methods enjoy safety, explainability, and stable generalization to new environments. However, when facing new environments, they still need a great deal of tuning, which often requires expert robotics knowledge~\cite{zheng2017ros, xiao2017uav}. 
Prior work has considered automated parameter tuning, e.g., finding trajectory optimization weights \cite{teso2019predictive} for the \textsc{dwa} planner \cite{fox1997dynamic}, or designing novel systems that can leverage gradient descent to match expert demonstrations \cite{bhardwaj2019differentiable}. Specifically, Xiao \emph{et al}~\cite{xiao2020appld} adopted black-box optimization to automatically map a robot's local observation to the optimal planner parameters via learning from human demonstration. While this technique can be applied to any parameter-based planner, it is not expected to generalize well in environments not seen in the demonstration. In contrast, \textsc{appli} only requires a few, short, local interventions when classical navigation does not perform well, instead of a demonstration of the full trajectory. \textsc{appli} also includes confidence estimation over candidate planner parameters during deployment in unseen environments. Notably, in the worst case our method reduces to the planner with default parameters, rather than a poorly chosen parameter set, and therefore enjoys better generalization. 

\subsection{Learning from Intervention} Due to the cost of providing full demonstrations, human intervention is a popular approach to providing minimal guidance for learning. It has been widely used in reinforcement learning \cite{saunders2017trial,prakash2019improving} and imitation learning \cite{goecks2019efficiently,kelly2019hg,zhang2016query,spencerlearning, kahn2021land}. Learning from intervention essentially focuses the agent on learning from its mistakes, thus improving the data efficiency and reducing the demonstration cost. In this work, we leverage the benefits of learning from intervention to enable robust robot navigation, and further categorize interventions based on the expert's estimation of the necessity of such interventions.

\subsection{Measuring Uncertainty in Deep Learning} Recent advances in deep learning have provided a family of tools for measuring the uncertainty in a deep model's prediction. There are mainly three types of approaches. (1) Bayesian Neural Networks (BNN) represent distributions over network weights and the prediction uncertainty is indirectly inferred via weight uncertainty \cite{kononenko1989bayesian}; (2) Deep Ensemble (DE) uses the outputs from multiple networks, each trained with partial data, as a Monte-Carlo estimator for uncertainty \cite{lee2019ensemble}. A specific example is Dropout learning\cite{gal2016dropout}; (3) Other methods train a single network with stationary weights but directly model the predictions in terms of a distribution. Evidential Deep Learning (EDL)\cite{sensoy2018evidential} is one particular method that models a discrete class of predictions with a Dirichlet distribution. We incorporate EDL into \textsc{appli} due to its simplicity and efficiency in terms of both time and space complexity, compared to methods from the other two approaches, which is essential for robot learning.

\section{APPROACH}
\label{sec::approach}

In this section, we introduce our method, \textsc{appli}, which has two novel features: (1) compared with Learning from Demonstration that requires demonstration of the whole task, \textsc{appli} only needs a few interventions in challenging scenarios where the default navigation system does not work well; (2) with a confidence measure on candidate parameters learned from interventions, our method knows when to switch back to the default parameters. This confidence measure enables \textsc{appli} to generalize well to unseen environments.

\subsection{Problem Definition}

We denote a classical parameterized navigation system as $G: \mathcal{X} \times \Theta \rightarrow \mathcal{A}$, where $\mathcal{X}$ is the state space of the robot (e.g. goal, sensor observations), $\Theta$ is the parameter space for $G$ (e.g. max speed, sampling rate, inflation radius), and $\mathcal{A}$ is the action space (e.g. linear and angular velocities). During deployment, the navigation system repeatedly estimates state $x$ and takes action $a$ calculated as $a = G(x; \bar{\theta})$. Typically, the default parameter set $\bar{\theta}$ is tuned by a human designer trying to achieve good performance in most environments. However, being good at everything often means being great at nothing: $\bar{\theta}$ usually exhibits suboptimal performance in some situations and may even fail (is unable to find feasible motions, or crashes into obstacles) in particularly challenging ones.

To mitigate this problem, a human can supervise the navigation system's performance at state $x$ by observing its action $a$ and judging whether (s)he should intervene. 
Here, we consider two types of interventions. A type A intervention is one in which the system performs so poorly that the human {\em must} intervene (e.g. imminent collision or a signal for help). A  type B intervention is one in which a human {\em might} intervene in order to improve otherwise suboptimal performance (e.g., driving too slowly in an open space). 
For the $i^\text{th}$ intervention, we assume that the human resets the robot to the position where the failure or suboptimal behavior first occurred and then gives a short teleoperated intervention $I_i=\{x_t, a_t\}_{t=1}^{T_i}$ of length $T_i$, where $x_{1:T_i}$ is the trajectory starting from the reset state induced by intervention actions $a_{1:T_i}$. As this short demonstration shows a cohesive navigation behavior in a specific segment of the environment (open space, narrow corridor, etc), we refer to the segment as a \textit{context} $c_i$ and denote the space of contexts as $\mathcal{C}$. Given $N$ interventions $I_{1:N}$, \textsc{appli} finds (1) a mapping $M: \mathcal{C} \rightarrow \Theta$ that determines the parameter set $\theta_i$ for each intervention context $c_i$, and (2) a parameterized predictor $B_\phi: \mathcal{X} \rightarrow \mathcal{C}$ that determines to which context (if any) the current state $x$ belongs.

\subsection{Parameter Learning}

After collecting a set of $N$ interventions $I_{1:N}$, for each $I_i$, we learn a set of navigation parameters $\theta_i$ that can best imitate the demonstrated correction. To find such parameters, we use the same training procedure as in the approach by Xiao \emph{et al}~\cite{xiao2020appld}, i.e., we use Behavior Cloning to minimize the difference between the actions from the human and those generated by the navigation system with new parameters $\theta_i$. To be specific, 

\begin{equation}
    \theta_i = \argmin_\theta \sum_{(x,a) \in I_i} \| a - G(x; \theta) \|_\lambda,
    \label{eq:bc_loss}
\end{equation}
where $\| d \|_\lambda=\sum \lambda_i d_i^2$ is the norm of the action difference with $\lambda$  weighting the different action dimensions (in our case, linear and angular velocity, $v$ and $\omega$). The loss in Eqn. (\ref{eq:bc_loss}) is minimized with a black-box optimizer, such as CMA-ES~\cite{hansen2003reducing}. After identifying parameters in each context, the mapping $M$ is simply $M(i) = \theta_i$.

\subsection{Confidence-Based Context Prediction}
So far, we have described how to learn multiple parameter sets $\theta_{1:N}$ from 
human interventions $I_{1:N}$ in contexts $c_{1:N}$. In order to select the correct parameters at deployment time, we must also determine if the current state $x_t$ falls into any one of the collected intervention contexts $c_i$. If such a determination can be made, then we direct the robot to use the parameter set $\theta_i$ to avoid making the same mistake as before. If it cannot be determined that $x_t$ belongs to a particular intervention context, then we direct the robot to use the default parameters $\bar{\theta}$, as they are optimized for most cases and are expected to generalize better than any parameter set learned for a specific scenario. In our system, the determination above is made using a predictor, $B_\phi$. To train this predictor, we first use the collected interventions to build a dataset, $\{\{x_t,c_i\}_{t=1}^{T_i}\}_{i=1}^N$, and train an intermediate classifier $f_\phi(x)$ with parameter set $\phi$ using the Evidential Deep Learning method (EDL) \cite{sensoy2018evidential}. A feature of EDL is that it supplies both a predicted label prediction and a confidence in that prediction $u_i \in (0, 1]$, i.e.,
\begin{equation}
    f_\phi(x_i) = (c_i, u_i).
\end{equation}
After training $f_\phi$ and during deployment, we can build a confidence-based classifier $g_\phi$ as 
\begin{equation}
    g_\phi(x_i) = c_i \mathds{1}(u_i \ge \epsilon_u),
\end{equation}
where $\epsilon_u$ is the threshold on confidence and $\mathds{1}$ is the indicator function. For state $x_i$, $g_\phi$ determine its context from $N+1$ contexts ($N$ intervention contexts and 1 default context). If $u_i \ge \epsilon_u$, it suggests the classifier $f_\phi$ is confident and $g_\phi$ predicts $c_i$. Otherwise, when $f_\phi$ is unsure about its prediction, $c_i \mathds{1}\{u_i \ge \epsilon_u \} = 0$. In this case, $g_\phi$ believes the current state $x_i$ is not similar to any intervention context and instead classifies $x_i$ as the default context. For this default context labeled as $c_i=0$, navigation utilizes the default navigation parameters $\bar{\theta}$ (i.e., we set $M(0) = \bar{\theta}$). 

Then we define our context predictor $B_\phi$ as:
\begin{equation}
    B_\phi(x_t) = \text{mode}( \left\{ g_\phi(x_i) \right\}_{i = t - w + 1}^t).
    \label{eq:context_predictor}
\end{equation}
To avoid a context estimation $c_t$ that changes frequently (e.g. caused by $g_\phi$'s wrong classifications), $B_\phi$ acts as a mode filter with window length $w$ and chooses the context $c_t$ that the majority of classifications agree with over the past $w$ time steps.

\subsection{\textsc{appli}}

Putting together all the components presented above, the entire \textsc{appli} pipeline is summarized in Alg. \ref{alg:APPLI}. In the training stage, it collects $N$ interventions from a human supervisor (line 1), and then learns corresponding navigation parameters $\theta_{1:N}$ i.e., the mapping $M$ (lines 2-3) and a context predictor $B_\phi$ (lines 5-6). During deployment, we use $M(B_\phi(x_t))$ to select the parameters for the navigation system at time $t$ (lines 8-10).

\begin{algorithm2e}
\SetAlgoNoLine
\caption{\strut \textsc{appli}}
\label{alg:APPLI}
\textbf{Training} \\
\KwIn{human interventions $I_{1:N} = \{\{x_t,a_t\}_{t=1}^{T_i}\}_{i=1}^N$, navigation system $G$, parameter space $\Theta$.}
\For{$i = 1, \dots, N$}{
    find parameter $\theta_i$ for context $i$ using Eqn. (\ref{eq:bc_loss}).
}
train the context classifier $f_\phi$ on $\{\{x_t,c_i\}_{t=1}^{T_i}\}_{i=1}^N$.\\
build mapping $M(i)=\theta_i$ and context predictor $B_\phi(x)$.

\textbf{Deployment} \\
\KwIn{navigation system $G$, parameter mapping $M$, context predictor $B_\phi(x)$, confidence threshold $\epsilon_u$, fallback parameters $\bar{\theta}$.}
\For{$t = 1, \dots$}{
	identify the current context $c_t = B_\phi(x_t)$ with confidence threshold $\epsilon_u$.\\
	navigate with $G(x_t, M(c_t))$.
}
\end{algorithm2e}
\section{EXPERIMENTS}
\label{sec::experiments}
In our experiments, we aim to show that \textsc{appli} can improve navigation performance by learning from only a few interventions and, with the confidence measurement, that the overall system can generalize well to unseen environments. We apply \textsc{appli} on a ClearPath Jackal ground robot in a physical obstacle course. Navigation performance learned through \textsc{appli} is then tested both in the same training environment, and also in another unseen physical test course. Furthermore, to investigate generalizability, we test the learned systems on a benchmark suite of 300 unseen simulated navigation environments.

\subsection{\textsc{appli} Implementation}

Our Jackal is a differential-drive robot equipped with a Velodyne LiDAR that we use to compute a 720-dimensional planar laser scan with a 270$^\circ$ field of view. The robot uses the Robot Operating System \texttt{move\textunderscore base} navigation stack with Dijkstra's global planner and the default \textsc{dwa} local planner, which works in most situations, but fails or behaves suboptimally in others (see Fig. \ref{fig::appli}).

During data collection, one of the authors (the {\em intervener}) follows the robot through the test course and intervenes when necessary, reporting if the intervention is to drive the robot out of a failure case (Type A) or to correct a suboptimal behavior (Type B). The four interventions are shown in Fig. \ref{fig::appli}: before the two Type A interventions (shown in red), the default system (\textsc{dwa} with $\bar{\theta}$) fails to plan feasible motions and starts recovery behaviors (rotates in place and moves backward); before the two Type B interventions (shown in yellow), the robot drives unnecessarily slowly in a relatively open space and enters the narrow corridor with unsmooth motions. For every intervention, the intervener stops the robot, drives it back to where they deem the failure or suboptimal behavior to have begun, and then provides recorded teleoperation $I$ that avoids the problematic behavior. To compare the performance learned from interventions and learned from a full demonstration, we also collect extra demonstrations for those places where the default planner already works very well (denoted in green in Fig. \ref{fig::appli}). 

This set of interventions comprises the input $I_{1:N} = \{\{x_t,a_t\}_{t=1}^T\}_{i=1}^N$ to Alg. \ref{alg:APPLI}, where $x_t$ is all the sensory data fed into the \texttt{move\textunderscore base} stack, $G$, and $a_t$ is the linear and angular velocity ($v$ and $\omega$) from teleoperation. 
The default and learned parameters are shown in Tab. \ref{tab::jackal_parameters}, including those learned from Type A and B interventions (A1, A2 and B1, B2), and the extra demonstrations (D1, D2).  

\begin{table}[h]
  \caption{Default and Learned Planner Parameters: \newline \emph{max\_vel\_x } \textnormal{(v)}, \emph{max\_vel\_theta} \textnormal{(w)}, \emph{vx\_samples} \textnormal{(s)}, \emph{vtheta\_samples} \textnormal{(t)}, \emph{occdist\_scale} \textnormal{(o)}, \emph{pdist\_scale} \textnormal{(p)}, \emph{gdist\_scale \textnormal{(g)}}, \emph{inflation\_radius \textnormal{(i)}}}
  \label{tab::jackal_parameters}
  \centering
  \small
  \begin{tabular}{lrrrrrrrr}
    \toprule
                & v & w & s & t & o & p & g & i \\
    \midrule
    \textsc{def.}       & 0.50 & 1.57 &  6 & 20 & 0.10 & 0.75 & 1.00 & 0.30 \\
    \midrule
    A1    & 0.26 & 2.00 & 13 & 44 & 0.57 & 0.76 & 0.94 & 0.02 \\
    A2    & 0.22 & 0.87 & 13 & 31 & 0.30 & 0.36 & 0.71 & 0.30 \\
    \midrule
    B1    & 1.91 & 1.70 & 10 & 47 & 0.08 & 0.71 & 0.35 & 0.23 \\
    B1    & 0.72 & 0.73 & 19 & 59 & 0.62 & 1.00 & 0.32 & 0.24 \\
    \midrule
    D1    & 0.37 & 1.33 &  9 &  6 & 0.95 & 0.83 & 0.93 & 0.01 \\
    D2    & 0.31 & 1.05 & 17 & 20 & 0.45 & 0.61 & 0.22 & 0.23 \\
    \midrule
    \bottomrule
  \end{tabular}
\end{table}

\subsection{Physical Experiments}
After the training in Alg. \ref{alg:APPLI} with the collected interventions, we deploy the learned mapping $M$ and context predictor $B_\phi$ on the \texttt{move\textunderscore base} navigation stack $G$. We use a confidence threshold $\epsilon_u=0.8$. 

\begin{figure}
  \centering
  \includegraphics[width=\columnwidth]{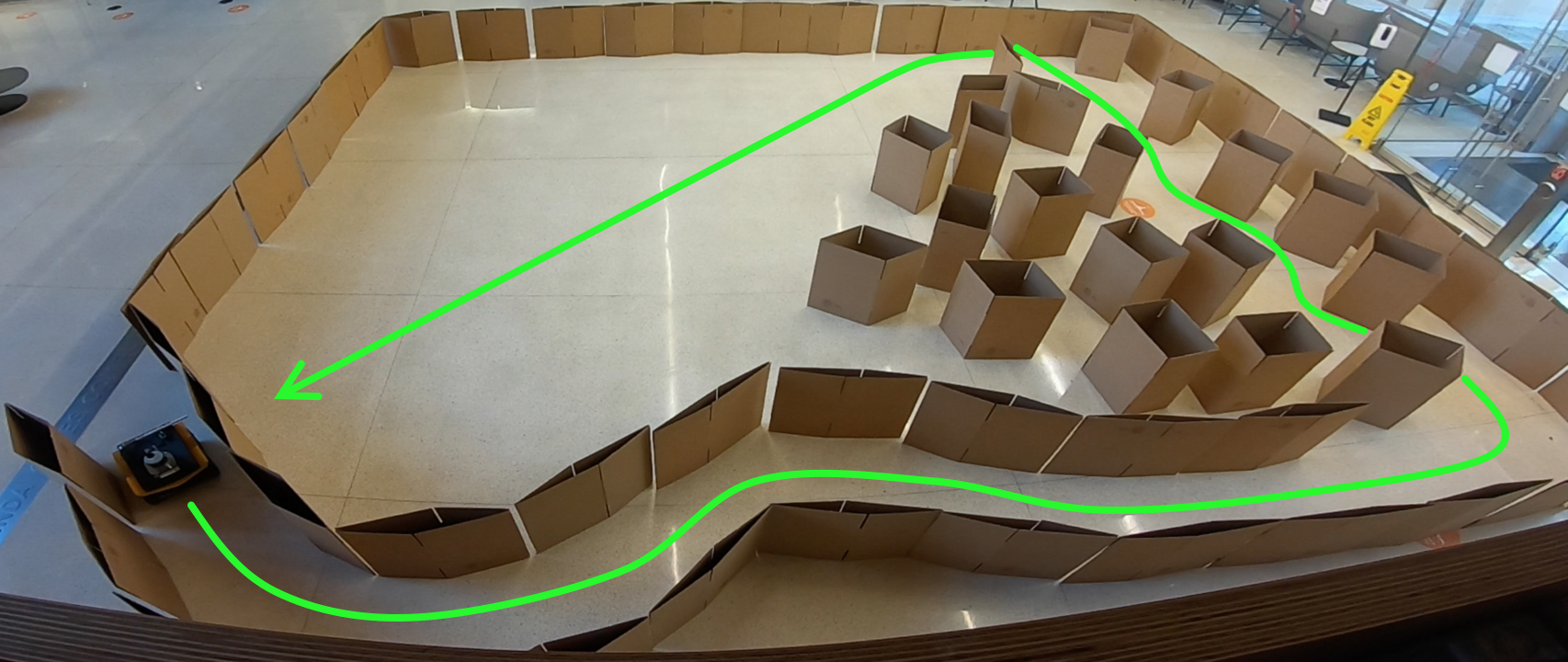}
  \caption{\textsc{appli} Running in an Unseen Physical Environment}
  \label{fig::unseen}
\end{figure}

We first deploy \textsc{appli} in the same training physical environment (Fig. \ref{fig::appli}). We compare the performance of the \textsc{dwa} planner with default parameters, \textsc{appli} learned only with Type A interventions, \textsc{appli} learned with Type A \emph{and} Type B interventions, and \textsc{appli} learned with a \emph{full demonstration} (which is basically the \textsc{appld} framework~\cite{xiao2020appld} enhanced by the confidence measure).
The motivation for the variation of \textsc{appli} learned only with Type A interventions is to study the effect of an unfocused or inexperienced human intervener. In this case, the human would still conduct all Type A interventions, as those mistakes are severe and easy to identify---some robots may even actively ask for help (e.g. by starting recovery behaviors). However, the human may fail to conduct Type B interventions as (s)he is not paying attention, or isn't equipped with the knowledge to identify suboptimal behaviors.
For all methods, we run five trials each and report the mean and standard deviation of the traversal time in Tab. \ref{tab::training_results}. If the robot gets stuck, we introduce a penalty value of 200 seconds. We also deploy the same sets of variants in an unseen physical environments (Fig. \ref{fig::unseen}) and report the results in Tab. \ref{tab::unseen_results}. 

\begin{table}
\centering
\caption{Traversal Time in Training Environment}
\begin{tabular}{cccc}
\toprule
Default & Type A & Type A+B & Full Demo \\ 
\midrule
134.0$\pm$60.6s & 77.4$\pm$2.8s  & 70.6$\pm$3.2s & 78.0$\pm$2.7s \\
\bottomrule
\end{tabular}
\label{tab::training_results}
\end{table}

\begin{table}
\centering
\caption{Traversal Time in Unseen Environment}
\begin{tabular}{cccc}
\toprule
Default & Type A & Type A+B & Full Demo\\ 
\midrule
109.2$\pm$50.8s & 71.0$\pm$0.7s & 59.0$\pm$0.7s & 62.0$\pm$2.0s\\
\bottomrule
\end{tabular}
\label{tab::unseen_results}
\end{table}

For both training and unseen environments, Type A interventions alone significantly improve upon the default parameters, by correcting all recovery behaviors such as rotating in place or driving backwards, and eliminating all failure cases. Adding Type B interventions further reduces traversal time, since the robot learns to speed up in relatively open spaces and to execute smooth motion when the tightness of the surrounding obstacles changes. All the interventions are able to improve navigation in both training and unseen environments, suggesting \textsc{appli}'s generalizability. Surprisingly, in both environments, \textsc{appli} learned from only Type A and Type B interventions can even outperform \textsc{appli} learned from an entire demonstration. One possible reason for this better performance from fewer human interactions is the additional human demonstrations may be suboptimal, especially since they are collected in places where the default navigation system was already deemed to have performed well. 
For example, in the full demonstration, we find the human intervener is more conservative than the default navigation system and drives slowly in some places. Hence, learning from these suboptimal behaviors introduces suboptimal parameters and consequently worse performance in contexts similar to that intervention.

\subsection{Simulated Experiments}

\begin{table*}[ht]
\vspace{0.25cm}
\centering
\caption{Percentage of Simulation Environments that Method 1 is Significantly Worse than Method 2 in Terms of Traversal Time\\
\scriptsize (Methods are listed in order of increasing performances. Results mentioned in experiment analysis are bold for better identification)}
\begin{tabular}{c l c c c c c c c}
\toprule
  & & \multicolumn{7}{c}{Method 2} \\
  & & \textsc{appli} (A) & \textsc{dwa} & \textsc{appli} (A+c) & \textsc{appli} (A+B+D+c) & \textsc{appli} (A+B+D) & \textsc{appli} (A+B+c) & \textsc{appli} (A+B)\\ 
\midrule
\multirow{7}{*}{\shortstack{Method\\1}} & \textsc{appli} (A) & 0   & 50    & \textbf{53}    & 62    & 63    & 68    & 66 \\
& \textsc{dwa}           & 10   & 0    & 6    & \textbf{33}    & 40    & \textbf{44}    & 47 \\
& \textsc{appli} (A+c)   & 6   & 4    & 0    & 31    & 37    & 45    & 45 \\
& \textsc{appli} (A+B+D+c) & 5   & 7    & 11    & 0    & 25    & \textbf{31}    & 33 \\
& \textsc{appli} (A+B+D) & 5   & 7    & 7    & 10    & 0    & \textbf{21}    & 21 \\
& \textsc{appli} (A+B+c) & 3   & 3    & 4    & 3    & 5    & 0   & 9 \\
& \textsc{appli} (A+B)   & 2   & 5    & 5    & 6    & 4    & 6   & 0 \\
\bottomrule
\end{tabular}
\label{tab::similation_results}
\end{table*}

To further test \textsc{appli}'s generalizability 
to unseen environments, we test our method with multiple variations and compare them with \textsc{dwa} on the Benchmark for Autonomous Robot Navigation (BARN) dataset~\cite{perille2020benchmarking}. The benchmark dataset consists of 300 simulated navigation environments generated using Cellular Automata, ranging from easy ones with a lot of open spaces to challenging ones where the robot needs to get through dense obstacles. Navigation trials in three example environments with low, medium, and high difficulty levels are shown in Fig. \ref{fig::simulated_envs}. Using the same training data collected from the physical environment shown in Fig.~\ref{fig::appli}, we test the following seven variants: 

\begin{figure}
  \centering
  \includegraphics[width=\columnwidth]{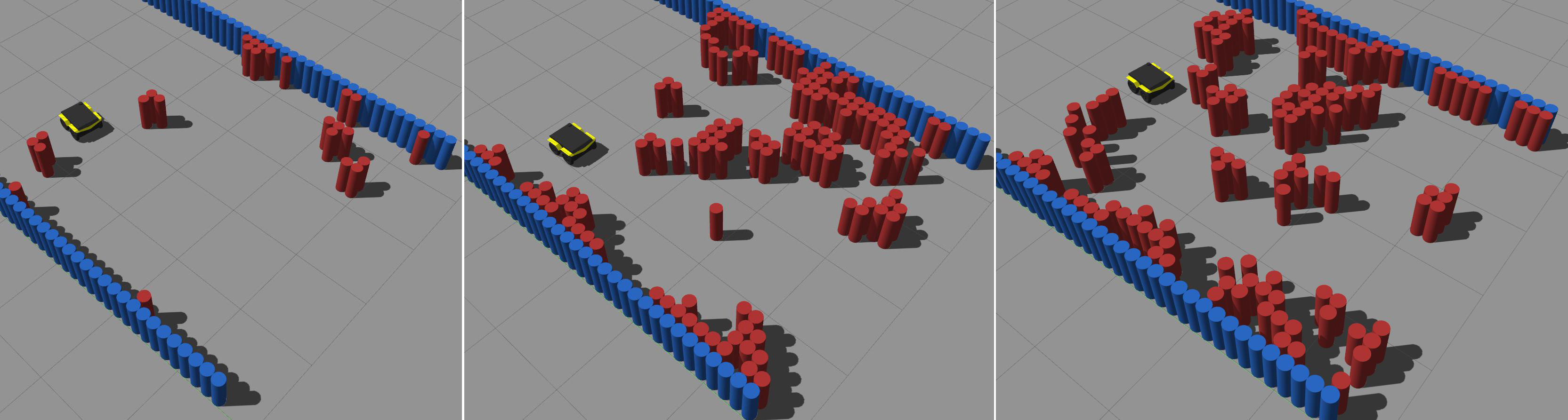}
  \caption{Navigation Trials in Example Environments with Low, Medium, and High Difficulty Levels}
  \label{fig::simulated_envs}
\end{figure}

\begin{figure}
  \centering
  \includegraphics[width=\columnwidth]{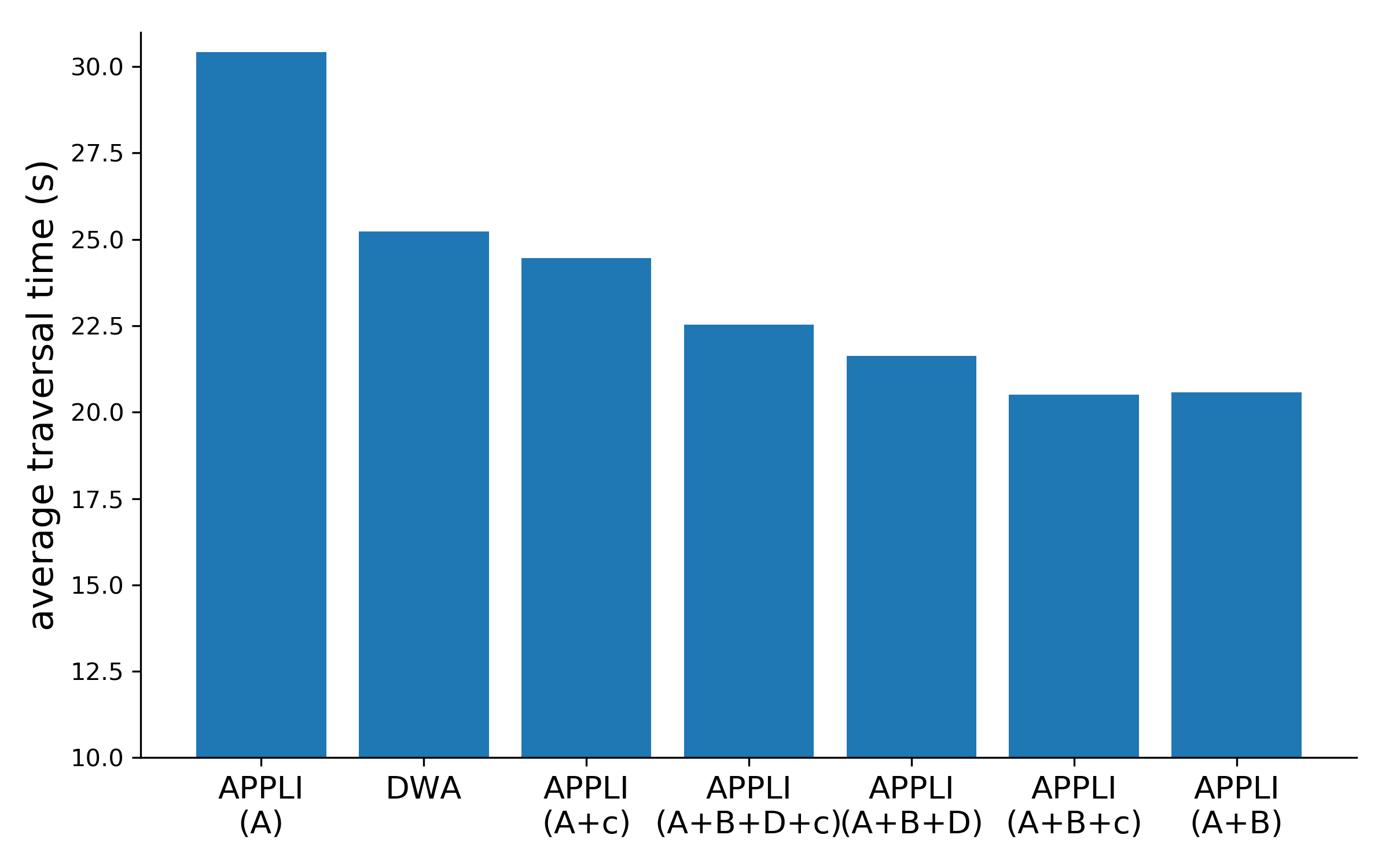}
  \caption{Average Performance in 300 Simulation Environments under 12 Different Runs}
  \label{fig::simulated_mean}
\end{figure}

\begin{itemize}
    \item \textsc{appli} learned from \emph{Type A and B} inventions \emph{with} confidence measure, denoted as \textsc{appli} (A+B+c).
    \item \textsc{appli} learned from \emph{Type A and B} inventions \emph{without} confidence measure, i.e., \textsc{appli} (A+B).
    \item \textsc{appli} learned from only \emph{Type A} interventions \emph{with} confidence measure, i.e., \textsc{appli} (A+c).
    \item \textsc{appli} learned from only \emph{Type A} interventions \emph{without} confidence measure, i.e., \textsc{appli} (A).
    \item \textsc{appli}  learned from \emph{full demonstration} \emph{with} confidence measure, i.e., \textsc{appli} (A+B+D+c).
    \item \textsc{appli}  learned from \emph{full demonstration} \emph{without} confidence measure, i.e., \textsc{appli} (A+B+D).
    \item the \textsc{dwa} planner with default parameters.
\end{itemize}

Testing these variations aims at studying the effect of learning from different modes of interventions caused by different degrees of human attention and experience levels, i.e. imperative interventions (A), optional interventions (A + B), and a full demonstration (A + B + D). They also provide an ablation study for the confidence measure in the EDL context classifier $f_\phi$: when deployed without the confidence measure, the robot has to choose among the parameters learned from interventions and never uses the default parameters. %

For each method in each simulation environment, we measure the traversal time for 12 different runs (the run is terminated after 50s if the robot gets stuck), resulting in 25200 total navigation trials. The average traversal time for each method in all simulation environments are shown in order of increasing performance in Fig. \ref{fig::simulated_mean}. We then conduct a pair-wise t-test for all methods in order to compute the percentage of environments in which one method (denoted as Method 1) is significantly worse ($p<0.05$) than another (denoted as Method 2). For better illustration, we also reorder the method by their performance and show the pairwise comparison in Tab. \ref{tab::similation_results}.

\textsc{appli} (A+B+c) and \textsc{appli} (A+B+D+c) outperform \textsc{dwa}: they are significantly better in 44\% and 33\% of environments respectively and significantly worse in only 3\% and 7\% of environments than \textsc{dwa}.
However, \textsc{appli} (A+c) is only significantly better than \textsc{dwa} 6\% of the time, which suggests that even though type B inventions only correct suboptimal performances, they are crucial for performance improvement. In detail, as the robot learns to go through narrow passages from Type A interventions, the first Type B intervention further teaches the robot to drive fast in safe open spaces, significantly reducing the traversal time for simulation environments whose beginnings and ends are relatively open. Meanwhile, from the second Type B intervention, the robot learns how to take sharp turns despite constrained surroundings.

In terms of the effect of confidence, 
\textsc{appli} (A) only selects parameters learned from 2 Type A inventions and never uses the default parameters even when they are more appropriate. Removing confidence greatly harms its performance, making it significantly worse than \textsc{appli} (A+c) in 53\% of environments. 
However, \textsc{appli} (A+B+c) and \textsc{appli} (A+B+D+c), which use more interventions or even the full demonstration to train the parameter mapping $M$ and context predictor $B_\phi$ are more confident about their predictions most of the time. As a result, removing confidence in the context predictor doesn't result in a significant difference.

Lastly, a conterintuitive, but similar result as in the physical experiments is that compared with \textsc{appli} (A+B+D+c) which uses the full demonstration, \textsc{appli} (A+B+c) learned from only Type A and B interventions achieves superior performance by being significantly better than \textsc{appli} (A+B+D+c) and \textsc{appli} (A+B+D) in 31\% and 21\% of the environments,  respectively. 
However, similar to the discussions about physical experiments, unnecessary human demonstrations are most likely suboptimal. In this sense, \textsc{appli} not only reduces the required human interactions from a full demonstration to only a few interventions, but also reduces the chance of performance degradation caused by suboptimal demonstrations.

\section{CONCLUSIONS}
\label{sec::conclusions}

In this work, we introduce \textsc{appli}, \emph{Adaptive Planner Parameter Learning from Interventions}. In contrast to most existing end-to-end machine learning for navigation approaches, \textsc{appli} utilizes existing classical navigation systems and inherits all their benefits, such as safety and explainability. Furthermore, instead of requiring a full expert demonstration or random exploration based on trial-and-error, \textsc{appli} only needs a few interventions, where the default underlying navigation system fails or exhibit poor behavior. It also introduces a confidence measure to assure generalizability in unseen environments. We show \textsc{appli}'s improved performance in training and unseen physical environments. We further test \textsc{appli}'s generalizability with 25200 simulated navigation trials in 300 unseen environments. While we allow the intervener to start interventions by rewinding the robot navigation to a state before the failure or suboptimal behavior occurs, an interesting direction for future work is to further investigate interventions without ``rewinding'', i.e. where the intervener takes over from where the robot fails and drives it forward to a good state.

\section*{ACKNOWLEDGMENTS}
This work has taken place in the Learning Agents Research
Group (LARG) at the Artificial Intelligence Laboratory, The University
of Texas at Austin.  LARG research is supported in part by grants from
the National Science Foundation (CPS-1739964, IIS-1724157,
NRI-1925082), the Office of Naval Research (N00014-18-2243), Future of
Life Institute (RFP2-000), Army Research Office (W911NF-19-2-0333),
DARPA, Lockheed Martin, General Motors, and Bosch.  The views and
conclusions contained in this document are those of the authors alone.
Peter Stone serves as the Executive Director of Sony AI America and
receives financial compensation for this work.  The terms of this
arrangement have been reviewed and approved by the University of Texas
at Austin in accordance with its policy on objectivity in research.

\bibliographystyle{IEEEtran}
\bibliography{IEEEabrv,references}

\begin{thebibliography}{10}
\providecommand{\url}[1]{#1}
\csname url@samestyle\endcsname
\providecommand{\newblock}{\relax}
\providecommand{\bibinfo}[2]{#2}
\providecommand{\BIBentrySTDinterwordspacing}{\spaceskip=0pt\relax}
\providecommand{\BIBentryALTinterwordstretchfactor}{4}
\providecommand{\BIBentryALTinterwordspacing}{\spaceskip=\fontdimen2\font plus
\BIBentryALTinterwordstretchfactor\fontdimen3\font minus
  \fontdimen4\font\relax}
\providecommand{\BIBforeignlanguage}[2]{{%
\expandafter\ifx\csname l@#1\endcsname\relax
\typeout{** WARNING: IEEEtran.bst: No hyphenation pattern has been}%
\typeout{** loaded for the language `#1'. Using the pattern for}%
\typeout{** the default language instead.}%
\else
\language=\csname l@#1\endcsname
\fi
#2}}
\providecommand{\BIBdecl}{\relax}
\BIBdecl

\bibitem{quinlan1993elastic}
S.~Quinlan and O.~Khatib, ``Elastic bands: Connecting path planning and
  control,'' in \emph{[1993] Proceedings IEEE International Conference on
  Robotics and Automation}.\hskip 1em plus 0.5em minus 0.4em\relax IEEE, 1993,
  pp. 802--807.

\bibitem{fox1997dynamic}
D.~Fox, W.~Burgard, and S.~Thrun, ``The dynamic window approach to collision
  avoidance,'' \emph{IEEE Robotics \& Automation Magazine}, vol.~4, no.~1, pp.
  23--33, 1997.

\bibitem{xiao2020toward}
X.~Xiao, B.~Liu, G.~Warnell, and P.~Stone, ``Toward agile maneuvers in highly
  constrained spaces: Learning from hallucination,'' \emph{arXiv preprint
  arXiv:2007.14479}, 2020.

\bibitem{liu2020lifelong}
B.~Liu, X.~Xiao, and P.~Stone, ``Lifelong navigation,'' \emph{arXiv preprint
  arXiv:2007.14486}, 2020.

\bibitem{xiao2020agile}
X.~Xiao, B.~Liu, and P.~Stone, ``Agile robot navigation through hallucinated
  learning and sober deployment,'' \emph{arXiv preprint arXiv:2010.08098},
  2020.

\bibitem{zheng2017ros}
K.~Zheng, ``Ros navigation tuning guide,'' \emph{arXiv preprint
  arXiv:1706.09068}, 2017.

\bibitem{xiao2017uav}
X.~Xiao, J.~Dufek, T.~Woodbury, and R.~Murphy, ``Uav assisted usv visual
  navigation for marine mass casualty incident response,'' in \emph{2017
  IEEE/RSJ International Conference on Intelligent Robots and Systems
  (IROS)}.\hskip 1em plus 0.5em minus 0.4em\relax IEEE, 2017, pp. 6105--6110.

\bibitem{teso2019predictive}
D.~Teso-Fz-Beto{\~n}o, E.~Zulueta, U.~Fernandez-Gamiz, A.~Saenz-Aguirre, and
  R.~Martinez, ``Predictive dynamic window approach development with artificial
  neural fuzzy inference improvement,'' \emph{Electronics}, vol.~8, no.~9, p.
  935, 2019.

\bibitem{bhardwaj2019differentiable}
M.~Bhardwaj, B.~Boots, and M.~Mukadam, ``Differentiable gaussian process motion
  planning,'' \emph{arXiv preprint arXiv:1907.09591}, 2019.

\bibitem{xiao2020appld}
X.~Xiao, B.~Liu, G.~Warnell, J.~Fink, and P.~Stone, ``Appld: Adaptive planner
  parameter learning from demonstration,'' \emph{IEEE Robotics and Automation
  Letters}, vol.~5, no.~3, pp. 4541--4547, 2020.

\bibitem{saunders2017trial}
W.~Saunders, G.~Sastry, A.~Stuhlmueller, and O.~Evans, ``Trial without error:
  Towards safe reinforcement learning via human intervention,'' \emph{arXiv
  preprint arXiv:1707.05173}, 2017.

\bibitem{prakash2019improving}
B.~Prakash, M.~Khatwani, N.~Waytowich, and T.~Mohsenin, ``Improving safety in
  reinforcement learning using model-based architectures and human
  intervention,'' \emph{arXiv preprint arXiv:1903.09328}, 2019.

\bibitem{goecks2019efficiently}
V.~G. Goecks, G.~M. Gremillion, V.~J. Lawhern, J.~Valasek, and N.~R. Waytowich,
  ``Efficiently combining human demonstrations and interventions for safe
  training of autonomous systems in real-time,'' in \emph{Proceedings of the
  AAAI Conference on Artificial Intelligence}, vol.~33, 2019, pp. 2462--2470.

\bibitem{kelly2019hg}
M.~Kelly, C.~Sidrane, K.~Driggs-Campbell, and M.~J. Kochenderfer, ``Hg-dagger:
  Interactive imitation learning with human experts,'' in \emph{2019
  International Conference on Robotics and Automation (ICRA)}.\hskip 1em plus
  0.5em minus 0.4em\relax IEEE, 2019, pp. 8077--8083.

\bibitem{zhang2016query}
J.~Zhang and K.~Cho, ``Query-efficient imitation learning for end-to-end
  autonomous driving,'' \emph{arXiv preprint arXiv:1605.06450}, 2016.

\bibitem{spencerlearning}
J.~Spencer, S.~Choudhury, M.~Barnes, M.~Schmittle, M.~Chiang, P.~Ramadge, and
  S.~Srinivasa, ``Learning from interventions,'' in \emph{Robotics: Science and
  Systems (RSS)}, 2020.

\bibitem{kahn2021land}
G.~Kahn, P.~Abbeel, and S.~Levine, ``Land: Learning to navigate from
  disengagements,'' \emph{IEEE Robotics and Automation Letters}, 2021.

\bibitem{kononenko1989bayesian}
I.~Kononenko, ``Bayesian neural networks,'' \emph{Biological Cybernetics},
  vol.~61, no.~5, pp. 361--370, 1989.

\bibitem{lee2019ensemble}
K.~Lee, Z.~Wang, B.~Vlahov, H.~Brar, and E.~A. Theodorou, ``Ensemble bayesian
  decision making with redundant deep perceptual control policies,'' in
  \emph{2019 18th IEEE International Conference On Machine Learning And
  Applications (ICMLA)}.\hskip 1em plus 0.5em minus 0.4em\relax IEEE, 2019, pp.
  831--837.

\bibitem{gal2016dropout}
Y.~Gal and Z.~Ghahramani, ``Dropout as a bayesian approximation: Representing
  model uncertainty in deep learning,'' in \emph{international conference on
  machine learning}, 2016, pp. 1050--1059.

\bibitem{sensoy2018evidential}
M.~Sensoy, L.~Kaplan, and M.~Kandemir, ``Evidential deep learning to quantify
  classification uncertainty,'' in \emph{Advances in Neural Information
  Processing Systems}, 2018, pp. 3179--3189.

\bibitem{hansen2003reducing}
N.~Hansen, S.~D. M{\"u}ller, and P.~Koumoutsakos, ``Reducing the time
  complexity of the derandomized evolution strategy with covariance matrix
  adaptation (cma-es),'' \emph{Evolutionary computation}, vol.~11, no.~1, pp.
  1--18, 2003.

\bibitem{perille2020benchmarking}
D.~Perille, A.~Truong, X.~Xiao, and P.~Stone, ``Benchmarking metric ground
  navigation,'' \emph{arXiv preprint arXiv:2008.13315}, 2020.

\end{thebibliography}

\end{document}